\documentclass[USletter,10pt,english]{article}
\usepackage{setspace}
\usepackage[margin=1in]{geometry}
\usepackage[T1]{fontenc}
\usepackage{ebgaramond}
\usepackage{amsfonts}       
\usepackage{hyperref}
\usepackage{algorithm}
\usepackage{algorithmicx}
\usepackage{amssymb}
\usepackage{amsmath}
\usepackage{graphicx,wrapfig,times,amsthm,bm}
\usepackage{subfig}
\usepackage{threeparttable}
\usepackage[wide]{sidecap}
\usepackage{array}
\usepackage{algorithm}
\usepackage{algpseudocode}
\usepackage{lipsum,babel}
\usepackage{color}
\usepackage{setspace}
\usepackage{psfrag,epsf}

\usepackage{amsbsy}
\usepackage{mathrsfs}
\usepackage{mathtools}
\usepackage{setspace,multirow, url,  subfig, bm, titlesec, textcomp, gensymb}

\def \vec {\mathrm{vec}}

\newcommand{\Mnorm}[2]{{\left\vert\kern-0.30ex\left\vert\kern-0.30ex\left\vert #1 
		\right\vert\kern-0.30ex\right\vert\kern-0.30ex\right\vert}}
\newcommand{\Opnorm}[3]{{\left\vert\kern-0.25ex\left\vert\kern-0.25ex\left\vert #1 
		\right\vert\kern-0.25ex\right\vert\kern-0.25ex\right\vert}_{#2 \to #3}}
\newcommand{\norm}[2]{{\left\vert\kern-0.30ex\left\vert #1 
		\right\vert\kern-0.30ex\right\vert}}

\newcommand{\innerproductminconstant}[1]{\psi_0}

\newif\ifarxiv
\arxivtrue
\arxivfalse

\newcommand{\sr}{\sigma^2_r}

\begin{document}
\ifarxiv
\doublespacing
\onecolumn
\else \fi

\title{Efficient Algorithms for Learning to Control Bandits \\with Unobserved Contexts} 
\author{Hongju Park and Mohamad Kazem Shirani Faradonbeh}%
\date{}
\maketitle


\begin{abstract}  
Contextual bandits are widely-used in the study of learning-based control policies for finite action spaces. While the problem is well-studied for bandits with perfectly observed context vectors, little is known about the case of imperfectly observed contexts. For this setting, existing approaches are inapplicable and new conceptual and technical frameworks are required.
We present an implementable posterior sampling algorithm for bandits with imperfect context observations and study its performance for learning optimal decisions. The provided numerical results relate the performance of the algorithm to different quantities of interest including the number of arms, dimensions, observation matrices, posterior rescaling factors, and signal-to-noise ratios. In general, the proposed algorithm exposes efficiency in learning from the noisy imperfect observations and taking actions accordingly. Enlightening understandings the analyses provide as well as interesting future directions it points to, are discussed as well.
\end{abstract}




\section{Introduction}
\label{sec:1}
Contextual bandits are commonly used for sequential decision-making with finitely many control actions. In this setting, available context observations can be utilized in a tractable way, thanks to the linearity of the relationship between the reward and the context vectors. The arms provide rewards depending on the contexts that represent their individual characteristics. The range of real-world applications is notably extensive, including personalized recommendations for Mobile Context-Aware Recommender Systems and  mobile-health interventions~\cite{bouneffouf2012contextual,tewari2017ads,nahum2018just}.

To get satisfactory performances in bandits, the exploration-exploitation trade-off must be addressed. The theoretical analysis of efficient policies for the multi-armed bandits goes back to algorithms that decide based on Upper-Confident-Bounds (UCB)~\cite{lai1985asymptotically}. In fact, UCB employs an optimistic approximate of the unknown reward based on the history of observations, to allow an appropriate degree of exploration. Further theoretical results for UCB in contextual bandits, as well as in other settings, are available in the literature~\cite{abe1999associative,auer2002using,abbasi2011improved,chu2011contextual,faradonbeh2020optimism}.

Posterior sampling is another ubiquitous reinforcement learning algorithm that effectively balances exploitation versus exploration. The idea of posterior sampling is suggested long time ago~\cite{thompson1933likelihood}, while its properties are adequately studied recently~\cite{strens2000bayesian,chapelle2011empirical,kaufmann2012thompson}. Theoretical regret bounds for bandits are established~\cite{agrawal2012analysis,agrawal2013thompson}, indicating that the posterior sampling algorithm is efficient, and can even outperform UCB. Further, posterior sampling exposes graceful performance in continuous action-spaces~\cite{faradonbeh2020adaptive} and large-scale problems~\cite{hu2019note}. 

In parallel, Greedy algorithms that correspond to certainty equivalence policies are studied for different problems. They can in general be inefficient for a class of multi-armed bandits, while are efficient in presence of a moderate degree of randomness~\cite{raghavan2020greedy,bastani2021mostly}. Moreover, for reinforcement learning problems in environments with memory, it is shown that (possibly randomized) Greedy policies perform efficiently. The existing literature varies along focusing on discrete and/or continuous time and spaces, including Markov decision processes \cite{raykar2014sequential,gopalan2015thompson,ouyang2017learning} 
and linear dynamical systems \cite{abbasi2011regret,faradonbeh2019applications,faradonbeh2020adaptive,faradonbeh2020input,faradonbeh2021efficient}. 

In several interesting situations that contextual bandits provide admissible models, context vectors are observed imperfectly and/or with measurement noise~\cite{kaelbling1998planning, spaan2012partially}. Further situations include where the entire feature vector is too costly to be inquired~\cite{bensoussan2004stochastic}. Moreover, in state space models, it is commonly assumed that the state vectors are imperfectly observed~\cite{roesser1975discrete,nagrath2006control,
kalman1960new,stratonovich1960application}. Thus, study of decision-making policies for bandits with imperfectly observable contexts is of interest. 
However, the available literature is notably sparse, and limited to empirical studies for combinatorial algorithms  \cite{bouneffouf2017context}, asymptotic analyses for noisy observations of contexts~\cite{park2021analysis}, and ridge regression in presence of side information about the parameter~\cite{tennenholtz2021bandits}.

This study aims to contribute to the subject by studying efficient algorithms for contextual bandits under imperfect observations, in absence of additional side information. In the sequel, we extensively investigate a class of scalable posterior sampling reinforcement learning policies. We study effects of different exogenous and indigenous parameters on the performance of the proposed algorithm for learning the unknown true parameter that determines the rewards, as well as on the regret one needs to incur.

The list of findings this work presents includes the followings. First, the number of arms has negligible effects on the parameter estimation, while its growth increases the regret. Second, the extent to which the observation matrix captures the unknown parameter (see \eqref{eq:ld}), significantly changes both the estimation error and regret. Third, by fattening the posterior to strengthen exploration, the regret grows, while the learning accuracy does \emph{not} improve. Fourth, for larger signal-to-noise ratios (of both the observation and the reward function), the regret growth improves but the learning error does \emph{not}. Fifth, larger dimensions of observations help faster parameter estimations, but make the optimal policy harder to compete against, leading to larger regrets. In the subsequent sections, we extensively discuss the details of the above findings, as well as the rationales they rely on.

The remainder of this paper is organized as follows. In Section~\ref{sec:2}, we state the problem and summarize preliminary materials. The posterior sampling algorithm for contextual bandits with imperfect context observations is presented and explained in \ref{sec:3}, followed by numerical analyses of the performance in Section~\ref{sec:4}. Finally, concluding remarks and future directions are discussed in Section~\ref{sec:5}.

\paragraph*{Notation:} 
For a matrix $A \in \mathbb{C}^{p \times q}$, $A^\top$ denotes its transpose. For a vector $v \in \mathbb{C}^d$, we use the $\ell_2$ norm $\|v\| = \left(\sum_{i=1}^d |v_i|^2\right)^{1/2}$. For matrices, we use the operator norm; $\|A\| = \sup_{\|v\|=1}  \|Av\|$, and $\mathrm{tr}(A)$ denotes the trace of $A$. Further, $C(A)$ and $C(A)^\perp$ denote the column-space of the matrix $A$ and its orthogonal space, respectively. Finally, $P_{C(A)}$ is the projection operator onto $C(A)$.

\section{Problem statement}
\label{sec:2}

In this section, we discuss the contextual bandit problem that multidimensional contexts corresponding to the arms are not available for decision-making. Instead, transformations of the context vectors with some observation noise are provided to the decision-maker to learn the optimal arm from. Technically, there are $N$ arms, each of which has its own $d_x$-dimensional context, denoted by $x_i(t)$ for arm $i \in \left\{1,\cdots, N \right\}$. The mechanism that generates context vectors over time consists of the normal distribution $N(0_{d_x},\Sigma_x)$. For the ease of presentation, the setting is assumed homoscedastic such that the covariance matrix $\Sigma_x$ is the same for all arms. However, extensions to heteroscedastic settings are similar. 

The $d_y$ dimensional observation $y_i(t)$ is given by
\begin{eqnarray}
y_i(t) = Ax_i(t) + \varepsilon_{y_i(t)}(t)\label{eq:obmod},
\end{eqnarray}
where $A$ is a $(d_y \times d_x)$ matrix capturing the relationship between $x_i(t)$ and $y_i(t)$ and $\varepsilon_{y_i(t)}(t) \sim N(0_{d_y},\Sigma_{y})$ is the measurement noise. So, $\Sigma_y$ is the conditional covariance matrix of $y_i(t)$ given $x_i(t)$. 

At each time $t$, the agent chooses an arm $a(t)$ and receives the reward $r_{a(t)}(t) = x_{a(t)}(t)^\top \mu_* + \varepsilon_{r_{a(t)}}(t)$. Above, $\mu_*$ is the unknown parameter that determines the expected reward and $\varepsilon_{r_{a(t)}}(t) \sim N(0,\sigma^2_r)$ is the reward noise. 

We focus on reinforcement learning policies for selecting an arm at every time step, to maximize the total reward in a long-term interaction with the environment. The presented policies utilize the collected data as time goes by, which consists of the observations, the actions taken to date, and the resulting rewards. 

For the observations attained through the model \eqref{eq:obmod}, the matrix $A$ determines the extent to which $y_i(t)$ can give information about $\mu_*$. Constrained by the above estimability restrictions of the observation model, we employ procedures that choose the arm maximizing $ \mathbb{E}\left[r_{i}(t) |y_i(t)\right] = y_{i}(t)^\top \eta_*$, instead of $x_i(t)^\top \mu_*$, where $\eta_*$ is the transformed parameter defined based on the $(d_x \times d_y)$ matrix $D=(A^\top \Sigma^{-1}_y A + \Sigma^{-1}_x)^{-1} A^\top \Sigma^{-1}_y$, according to
\begin{eqnarray}
\eta_* = D^\top \mu_*. \label{eq:eta}
\end{eqnarray}

So, define $\ell_{D,\mu_*}$ to quantify the estimability of $\mu_*$:
\begin{eqnarray}
\ell_{D,\mu_*}=\frac{\|P_{C(D)}\mu_*\|}{\|\mu_*\|}.\label{eq:ld}
\end{eqnarray}
In fact, $\ell_{D,\mu_*}$ represents the proportion of the parameter $\mu_*$ that can be learned by using the observations together with the rewards. To see the rationale of the above formula and its effect on the performance of reinforcement learning algorithms, consider the following special cases. If $\mu_* \in C(D)$, which means $\ell_{D,\mu_*}=1$, then $\mu_*$ is totally estimable in the sense that an estimator converges to $\mu_*$ as data accumulate. Otherwise, if $\mu_* \in C(D)^\perp$, then $\mu_*$ is not estimable at all. In the latter case, an estimator for $\mu_*$ cannot converge to $\mu_*$, even with infinite data. However, regardless of the value of the ratio $\ell_{D,\mu_*}$, the expected reward given the observation $y_i(t)$ depends on the transformed parameter $\eta_*$, which is by definition estimable.  
Thus, the optimal arm at time $t$ becomes
\begin{eqnarray}
a^*(t) = \underset{1 \leq i \leq N}{\arg\max}~ \mathbb{E}\left[r_{i}(t) |y_i(t)\right] =\underset{1 \leq i \leq N}{\arg\max}~ y_i(t)^{\top} \eta_* \label{eq:a*t}.
\end{eqnarray}
Since uncertainties prevent the adaptive decision-maker from repeatedly choosing the optimal arm(s), a quantity is needed to assess the algorithms. Regret is a common performance measure in reinforcement learning for reflecting the sub-optimalities incurred due to uncertainty about the true parameter $\mu_*$. It is defined as the cumulative reward difference between the optimal and the chosen arms:
\begin{eqnarray}
\mathrm{Regret}(T) &=& \sum_{t=1}^T(r_{a^*(t)}(t)  - r_{a(t)}(t)),\label{eq:reg}
\end{eqnarray}



where $a(t)$ is the arm chosen at time $t$. Note that the regret above is stochastic and provides a worst-case criteria, and so is more general than the expected regret that reflects average-case sub-optimalities. 


\section{Reinforcement Learning Algorithm}
\label{sec:3}

In this section, we introduce a reinforcement learning algorithm for contextual bandits with imperfectly observed contexts. The algorithm uses posterior sampling. The goal of the algorithm is to maximize the cumulative reward given the observations $\{y_i(t)\}_{1\leq i \leq N, 1\leq t \leq T}$, because the contexts are not observed. To incorporate this, the algorithm is designed aiming to minimize the conditional gap
\begin{eqnarray}
\mathbb{E}\left[\left.r_{a^*(t)}(t)  - r_{a(t)}(t) \right|\{y_i(t)\}_{1\leq i \leq N} \right]. \label{eq:reg2}
\end{eqnarray}

That is, to minimize the regret in \eqref{eq:reg}, at each time $t$, the proposed algorithm tries to choose the arm $a(t)$ that minimizes the conditional expectation in \eqref{eq:reg2}. To express in terms of $y_i(t)$ and $\mu_*$, we use the conditional distribution of $r_i(t)$ given $y_i(t)$;
\begin{eqnarray}
    N \left( y_i(t)^\top D^\top \mu_* ,~\mu_*^\top(A^{\top} \Sigma_y^{-1} A + \Sigma_x^{-1})^{-1}  \mu_* + \sigma^2_r \right)\label{eq:ldry},
\end{eqnarray}
where the matrix $D$ that reflects the average effect of $y_i(t)$ on $r_i(t)$ is defined in Section \ref{sec:2}. 
Further, based on the estimability of $\mu_*$ (discussed in Section \ref{sec:2}), the algorithm estimates $\eta_*$ in \eqref{eq:eta}. To that end, it employs the rescaled posterior distribution
\begin{eqnarray}
N(\widehat{\eta}(t), c_B B(t)^{-1}),\label{eq:pos}
\end{eqnarray}
where $\widehat{\eta}(t)$ is the posterior mean and $B(t)$ is the inverse posterior covariance $B(t)$, given by
\begin{eqnarray}
B(t+1) &=& B(t) + y_{a(t)}(t) y_{a(t)}(t)^{\top}, \label{eq:B}\\
\widehat{\eta}(t+1) &=& B(t+1)^{-1}(B(t)\widehat{\eta}(t) + y_{a(t)}(t) r_{a(t)}(t)).\:\:\:\:~~\label{eq:etahat}
\end{eqnarray}
The initial values are $\widehat{\eta}(1) = 0_{d_y}$ and $B(1) = I_{d_y}$. Importantly, the constant $c_B$ is used for rescaling the posterior, so that larger $c_B$ leads to more exploration, and vice versa. Then, the posterior sampling algorithm chooses the arm $a(t)$ that

\begin{eqnarray}
a(t) =\underset{1 \leq i \leq N}{\arg\max}~ y_i(t)^{\top} \widetilde{\eta}(t),\label{eq:at}
\end{eqnarray}

where $\widetilde{\eta}(t)$ is a sample from the posterior in \eqref{eq:pos}.

The pseudo-code of the posterior sampling algorithm explained above is provided in Algorithm~\ref{algo}. At every time step, Algorithm~\ref{algo} chooses the above arm $a(t)$. Then, with the obtained observations $y_{a(t)}(t)$ and reward $r_{a(t)}(t)$, the algorithm updates the posterior distribution based on \eqref{eq:B} and \eqref{eq:etahat}. The case $c_B=0$ represents Greedy algorithm, where the posterior is concentrated at its mean, and so it holds that $\widetilde{\eta}(t) = \widehat{\eta}(t) $.

\begin{algorithm}[h] 
	\begin{algorithmic}[1]
		\State Set $B(1) = I_{d_y}$, $\widehat{\eta}(1) = \mathbf{0}_d$
		\For{$t = 1,2, \dots, $} 
		\State Sample $\widetilde{\eta}(t)$ from $N(\widehat{\eta}(t), c_B B(t)^{-1})$
		\State Select arm $a(t)= \underset{1 \leq i \leq N}{\arg\max}~ y_i(t)^{\top} \widetilde{\eta}(t)$
		\State Gain reward $r_{a(t)}(t) = x_{a(t)}(t)^\top \mu_* + \epsilon_{r_{a(t)}}(t)$
		\State Update $B(t+1)$ and $\widehat{\eta}(t+1)$ by \eqref{eq:B} and  \eqref{eq:etahat}
		\EndFor
	\end{algorithmic}
\caption{: Posterior sampling policy for bandits with imperfect context observations}  \label{algo}
\end{algorithm}



\begin{figure*}[h]
    \centering
    \psfrag{N=5~~~~}{\scriptsize$N=5$}
    \psfrag{N=10~~~~}{\scriptsize$N=10$}
    \psfrag{N=20~~~~}{\scriptsize$N=20$}
    \psfrag{N=50~~~~}{\scriptsize$N=50$}
    \psfrag{A~~~~~~~~~~~~~~~~~~~~~~~~~~~~~~~~~}{~~~~~~~~Average Estimation Error of $\widehat{\eta}(t)$ vs Time}
    \psfrag{N~~~~~~~~~~~~~~~~~~~~~~}{~~~~~~Normalized Regret vs Time}
    \psfrag{t}{$t$}
    \includegraphics[width=0.88\textwidth]{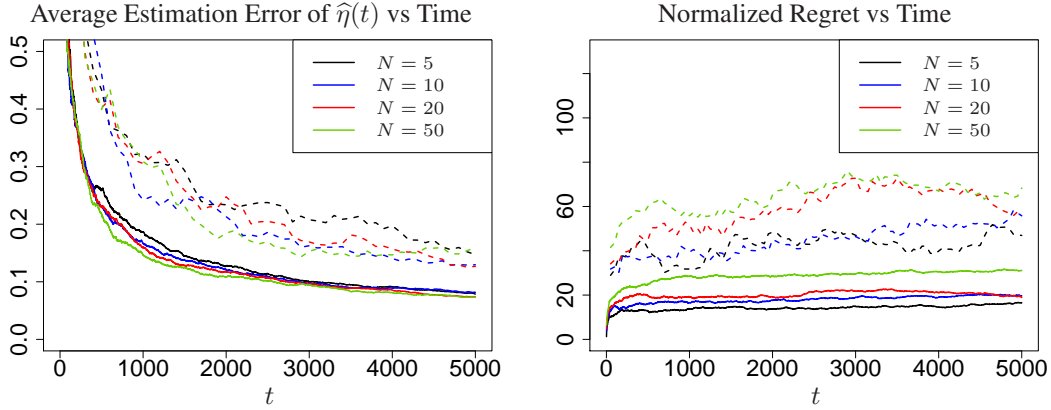}
    \caption{Plots of $\|\widehat{\eta}(t)-\eta_*\|$ and $\mathrm{Regret}(t)/\log t$ over time, for the different number of arms $N = 5,10,20,50$. The dashed (solid) lines represent the worst- (average-)case performance among $50$ scenarios.}
    
    
    \label{fig:n}
\end{figure*}

\begin{figure*}[h]
    \centering
    \psfrag{~~~~c=0.25}{\scriptsize$\ell_{D,\mu_*}=0.25$}
    \psfrag{~~~~c=0.5}{\scriptsize$\ell_{D,\mu_*}=0.5$}
    \psfrag{~~~~c=0.75}{\scriptsize$\ell_{D,\mu_*}=0.75$}
    \psfrag{~~~~c=1}{\scriptsize$\ell_{D,\mu_*}=1$}
    \psfrag{A1~~~~~~~~~~~~~~~~~~~~~~~~~~~~~~~~~~~~}{\scriptsize ~~Average Estimation Error of $\widehat{\eta}(t)$ vs Time}
    \psfrag{A2~~~~~~~~~~~~~~~~~~~~~~~~~~~~~~~~~~~~}{\scriptsize ~~Average Estimation Error of $\widehat{\mu}(t)$ vs Time}
    \psfrag{N~~~~~~~~~~~~~~~~~~~~~~~~}{\scriptsize Normalized Regret vs Time}
    \psfrag{1.1}{\scriptsize0}
    \psfrag{2.2}{\scriptsize1}
    \psfrag{3.3}{\scriptsize10}
    \psfrag{4.4}{\scriptsize100}
    \psfrag{cB}{$c_B$}
    \psfrag{t}{$t$}
    \includegraphics[width=0.96\textwidth]{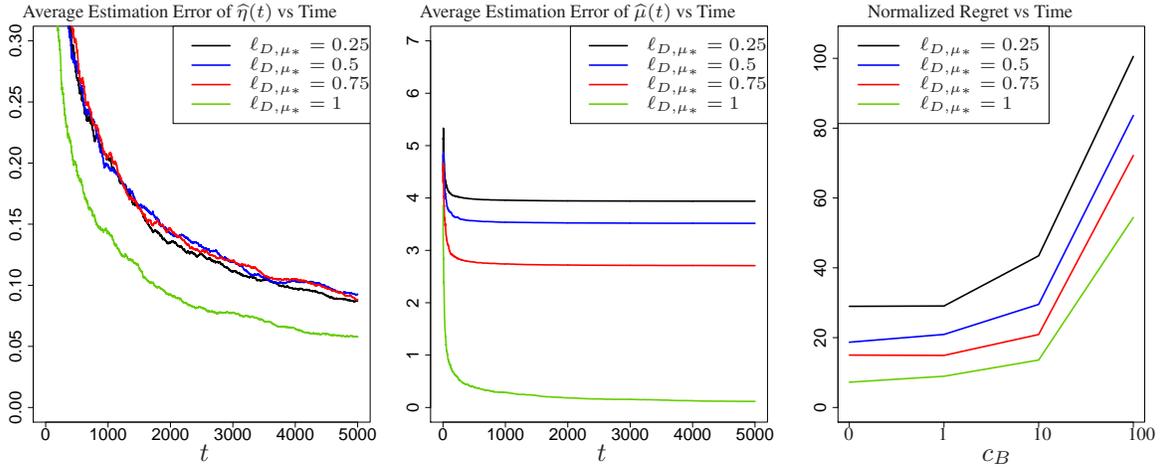}
    \caption{Plots of $\|\widehat{\eta}(t)-\eta_*\|$, $\|\widehat{\mu}(t)-\mu_*\|$ versus $t$, and $\mathrm{Regret}(5000)/\log 5000$ versus $c_B$, for different values of $\ell_{D,\mu_*}$.
    }
    \label{fig:mu}
\end{figure*}

\section{Analysis of Algorithm~\ref{algo}}

\label{sec:4}
For studying the presented algorithm, we consider various settings with different parameter setups. To understand how different parameters affect the estimation errors and the regret, we report the values of $\|\widehat{\eta}(t)-\eta_*\|$, $\|\widehat{\mu}(t)-\mu_*\|$, and $\mathrm{Regret}(t)/\log t$ for different observation matrices $A$, as well as different number of arms, dimensions, posterior rescaling constant $c_B$, and the signal-to-noise ratios $\boldsymbol{\mathrm{SNR}}_y$ and $\boldsymbol{\mathrm{SNR}}_r$, which will be defined later. Here, $\widehat{\mu}(t)$ is calculated based on the relationship $\widehat{\mu}(t) = (D B(t) D^\top )^{-1}D B(t) \widehat{\eta}(t)$, which is derived with the relationship between the posterior means of $\mu_*$ and $\eta_*$. Henceforth, when the exact values of the parameters are not explicitly mentioned, they are fixed to the following default values: 
\begin{eqnarray}
d_x=20,~~~~d_y=5,~~~~N=5,~~~~\sr = 1,~~~~\Sigma_x = I_{d_x},~~~~\Sigma_y = I_{d_y},~~~~c_B = 0.~~~~\nonumber
\end{eqnarray}
Further, the rows of $A$ are randomly generated and then adjusted to form an orthonormal basis. For each setting, $50$ scenarios are simulated.

Figure \ref{fig:n} provides the effect of the number of arms $N$ on the estimation errors and the regret. The left plot of Figure \ref{fig:n} shows that the number of arms does not affect the estimation. This is because, regardless of the number of the bandit arms, at every time step only the reward and the observation vector of the chosen arm $a(t)$ provide data for learning the unknown parameter. However, the right plot of Figure \ref{fig:n} illustrates that the regret bound grows as $N$ increases. This is intuitively expected since finding optimal arms becomes harder as $N$ grows.  

In Figure \ref{fig:n}, the dashed lines in the both graphs indicate the worse-case quantity at each time $t$, while the average-case quantities are shown by solid lines. Because the two types of the curves are not far apart, henceforth we only report the average-case analyses. 


Figure \ref{fig:mu} depicts how the matrix $A$ affects the estimation of $\eta_*$, $\mu_*$, as well as the regret. 
To study the effect of $A$ on estimability, $A$ is generated to let $\ell_{D,\mu_*}=0.25,~0.5,~0.75,~1$. In the leftmost plot in Figure \ref{fig:mu}, $\ell_{D,\mu_*}$ does not have a significant effect on the estimation of $\eta_*$, while significantly smaller errors occur $\ell_{D,\mu_*}=1$ thanks to the stable posterior covariance structure of $\eta_*$. In the middle plot, when $\ell_{D,\mu_*}$ is less than $1$, the estimation errors do not decay over time, while they do $\ell_{D,\mu_*} = 1$. Note that the latter phenomena is totally consistent with the definition of the estimability factor $\ell_{D,\mu_*}$.

In addition, the regret decreases as $\ell_{D,\mu_*}$ increases. The rightmost plot in Figure \ref{fig:mu} illustrates the relationship between the normalized regret and the posterior rescaling constant $c_B$, for different values of $\ell_{D,\mu_*}$. The regret curves indicate that Greedy algorithm (i.e., $c_B=0$) and small values of $c_B$ expose the best performance. Thus, additional exploration is not needed as the random context and output vectors provide sufficient explorations. The effect of $c_B$ on the estimation error is so negligible that it is not reported. 

\begin{figure*}[h]
    \centering
    \psfrag{SNR_r}{\scriptsize$\boldsymbol{\mathrm{SNR}}_r$}
    \psfrag{1.1}{\scriptsize$\frac{1}{4}$}
    \psfrag{2.2}{\scriptsize$\frac{1}{2}$}
    \psfrag{3.3}{\scriptsize$1$}
    \psfrag{4.4}{\scriptsize$2$}
    \psfrag{5.5}{\scriptsize$4$}
    \psfrag{SNR_y=4/1~~}{\scriptsize$\boldsymbol{\mathrm{SNR}}_y=4$}
    \psfrag{SNR_y=2/1~~}{\scriptsize$\boldsymbol{\mathrm{SNR}}_y=2$}
    \psfrag{SNR_y=1/1~~}{\scriptsize$\boldsymbol{\mathrm{SNR}}_y=1$}
    \psfrag{SNR_y=1/2~~}{\scriptsize$\boldsymbol{\mathrm{SNR}}_y=\frac{1}{2}$}
    \psfrag{SNR_y=1/4~~}{\scriptsize$\boldsymbol{\mathrm{SNR}}_y=\frac{1}{4}$}
    \psfrag{A1~~~~~~~~~~~~~~~~~~~~~~~~~~~~~~~~~~~~}{\scriptsize ~~~~~~~~Average Estimation Error of $\widehat{\eta}(t)$ vs $\boldsymbol{\mathrm{SNR}}_r$}
    \psfrag{A2~~~~~~~~~~~~~~~~~~~~~~~~~~~~~~~~~~~~}{\scriptsize ~~~~~~~~Average Estimation Error of $\widehat{\mu}(t)$ vs $\boldsymbol{\mathrm{SNR}}_r$}
    \psfrag{N~~~~~~~~~~~~~~~~~~~~~~~~}{\scriptsize ~~~~~Normalized Regret vs $\boldsymbol{\mathrm{SNR}}_r$}
    \psfrag{t}{$t$}
    \includegraphics[width=0.96\textwidth]{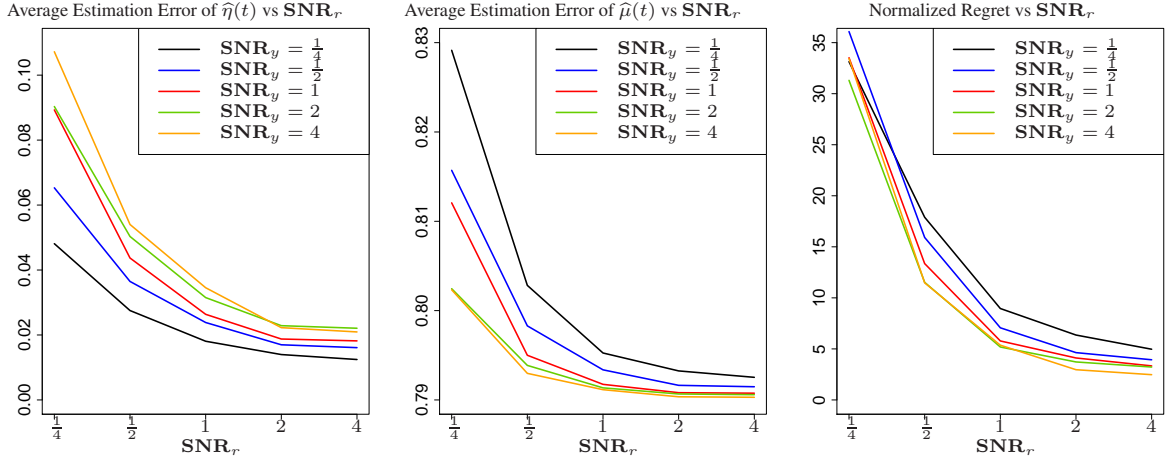}
    \caption{Plots of $\|\widehat{\eta}(T)-\eta_*\|$, $\|\widehat{\mu}(T)-\mu_*\|$ and $\mathrm{Regret}(T)/\log T$ at $T=5000$, versus to the signal-to-noise ratio of the reward function, for different signal strengths of the observation model. 
    }
    \label{fig:snr}
\end{figure*}

\begin{figure*}[h]
    \centering
    \psfrag{dy=5~~~~}{\scriptsize$d_y=5$}
    \psfrag{dy=20~~~~}{\scriptsize$d_y=20$}
    \psfrag{dy=50~~~~}{\scriptsize$d_y=50$}
    \psfrag{dy=100~~~~}{\scriptsize$d_y=100$}
    \psfrag{Error~~~~~~~~~~~~~~~~~~~~~~~}{$\mathbb{E}\left[\|\widehat{\mu}(t) - \mu_*\|\right]$}
    \psfrag{Regret(t)/log(t) vs Time~~~~~~~~~~}{~~~$\mathbb{E}\left[\mathrm{Regret}(t)\right]/\log(t)$ vs Time}
    \psfrag{t}{$t$}
    \psfrag{Estimation Error vs Time}{$\mathbb{E}\left[\|\widehat{\mu}(t) - \mu_*\|\right]$ vs Time}
    \psfrag{A2~~~~~~~~~~~~~~~~~~~~~~~~~~~~~~~~~~~~}{\scriptsize ~~Average Estimation Error of $\widehat{\mu}(t)$ vs Time}
    \psfrag{N~~~~~~~~~~~~~~~~~~~~~~~~}{\scriptsize Normalized Regret vs Time}
    \psfrag{C~~~~~~~~~~~~~~~~~~~~~~~~}{\scriptsize Cumulative Reward vs Time}
    \includegraphics[width=0.96\textwidth]{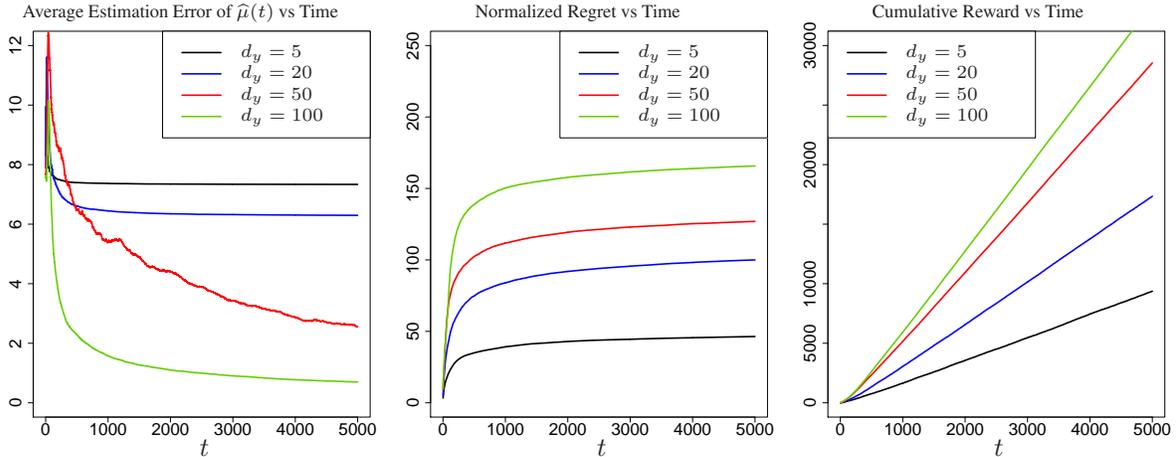}
    \caption{Plots of $\|\widehat{\mu}(t)-\mu_*\|$, $\mathrm{Regret}(t)/\log t$ and $\sum_{\tau=1}^t r_{a(\tau)}(\tau)$ over time, for the different dimensions of the observations; $d_y = 5,20,50,100$ and $d_x=50$.
    }
    \label{fig:dy}
\end{figure*}

Figure \ref{fig:snr} showcases the effects of the signal-to-noise ratios on the performances of the algorithm. Based on the reward function $r_i(t) = x_i(t)^\top \mu_* + \varepsilon_{r_i(t)}(t)$, the ratio for the reward is defined as:
\begin{eqnarray}
\boldsymbol{\mathrm{SNR}}_r = \frac{\mathbb{E}[(x_i(t)^\top \mu_*)^2]}{\sr} =  \frac{\mu_*^\top \Sigma_x\mu_* }{\sr}.
\end{eqnarray}
 Similarly, based on the observation model $y_i(t)=A x_i(t) + \varepsilon_{y_i(t)}(t)$, the ratio for the observations is
 
\begin{eqnarray}
\boldsymbol{\mathrm{SNR}}_y = \frac{\mathbb{E}[\|A x_i(t)\|^2]}{\mathbb{E}[\|\varepsilon_{y_i(t)}(t)\|^2]} = \frac{\mathrm{tr}(A \Sigma_x A^\top)}{\mathrm{tr}(\Sigma_y)}.
\end{eqnarray}
For ease of demonstration, we let $\Sigma_x=s_xI_{d_x}$ and $\Sigma_y = s_y I_{d_y}$, which yield to $\boldsymbol{\mathrm{SNR}}_y = {s_x}/{s_y}$, because $A$ sets to be orthonormal.
Generally, stronger signals lead to more accurate estimations and so better performances. However, in Figure \ref{fig:snr} we see that $\boldsymbol{\mathrm{SNR}}_y$ has effects on the estimations of $\eta_*$ and $\mu_*$ in opposite directions. Note that greater $\boldsymbol{\mathrm{SNR}}_r$ always helps both of the estimations. 
That is because the accuracy of learning $\eta_*$ depends on diversity of $y_i(t)$ vectors such that for smaller $\boldsymbol{\mathrm{SNR}}_y$, $y_i(t)$ exposes more random behavior and so the matrix $B(t)$ grows in all directions, leading to richer data for faster estimation.
However, noting the scale of graphs, especially the middle plot in Figure \ref{fig:snr}, this effect is very small, and for learning $\mu_*$, it is practically negligible. On the other hand, for the growth of regret, stronger signals reduce the sub-optimality, especially the effect of $\boldsymbol{\mathrm{SNR}}_r$ is visible in Figure~\ref{fig:snr}.


Figure \ref{fig:dy} illustrates the performance of Algorithm~\ref{algo} for different dimension values of observations. Here, $d_x$ is fixed at $50$ and $d_y$ is gradually changed from $5$ to $100$. For this setting, each row of $A$ is generated from the multivariate normal distribution and normalized, but the rows of $A$ are not orthogonal. Basically, as seen in the leftmost plot, a larger $d_y$ gives more information about the contexts, which in turn decreases the estimation error. 

In contrast to that, the regret curves in the middle plot show that a larger dimension of the observations has a negative effect on the performance of the algorithm. Note that according to the plot on the left, this is not caused by the error in learning the true parameter. Instead, it is due to the fact that larger $d_y$ lets the optimal policy \eqref{eq:a*t} estimate the unobserved contexts more accurately, such that competing against it becomes harder for the reinforcement learning policy of Algorithm~\ref{algo}. To see that better, observe the rightmost plot for cumulative rewards of the policy of Algorithm~\ref{algo} versus time, which shows that larger $d_y$ leads to significantly greater cumulative rewards.


\section{Conclusion and Future Work}
\label{sec:5}

We studied the posterior sampling reinforcement learning algorithm for contextual multi-armed bandits with imperfect context observations. Extensive numerical analyses are provided and the resulting performance of the presented algorithm for learning the unknown true parameters and for competing against the optimal policy are discussed. The relationships between the different quantities, including time horizon, number of arms, dimension of observations, measurement matrix, posterior rescaling factor, and signal-to-noise ratios are fully investigated. 


As an initiating study about imperfectly observed contextual bandits, this work spotlights several interesting directions for future research. In fact, it will be of remarkable interests to develop theoretical guarantees and provable performance results for scaling of the estimation error and regret with different quantities introduced in this paper. Further, tackling unknown observation models with an estimation procedure for the measurement matrix is an interesting extension of the current work. 

\bibliographystyle{IEEEtran}   
\bibliography{mybib}

\begin{thebibliography}{10}
\providecommand{\url}[1]{#1}
\csname url@samestyle\endcsname
\providecommand{\newblock}{\relax}
\providecommand{\bibinfo}[2]{#2}
\providecommand{\BIBentrySTDinterwordspacing}{\spaceskip=0pt\relax}
\providecommand{\BIBentryALTinterwordstretchfactor}{4}
\providecommand{\BIBentryALTinterwordspacing}{\spaceskip=\fontdimen2\font plus
\BIBentryALTinterwordstretchfactor\fontdimen3\font minus
  \fontdimen4\font\relax}
\providecommand{\BIBforeignlanguage}[2]{{%
\expandafter\ifx\csname l@#1\endcsname\relax
\typeout{** WARNING: IEEEtran.bst: No hyphenation pattern has been}%
\typeout{** loaded for the language `#1'. Using the pattern for}%
\typeout{** the default language instead.}%
\else
\language=\csname l@#1\endcsname
\fi
#2}}
\providecommand{\BIBdecl}{\relax}
\BIBdecl

\bibitem{bouneffouf2012contextual}
D.~Bouneffouf, A.~Bouzeghoub, and A.~L. Gan{\c{c}}arski, ``A contextual-bandit
  algorithm for mobile context-aware recommender system,'' in
  \emph{International conference on neural information processing}.\hskip 1em
  plus 0.5em minus 0.4em\relax Springer, 2012, pp. 324--331.

\bibitem{tewari2017ads}
A.~Tewari and S.~A. Murphy, ``From ads to interventions: Contextual bandits in
  mobile health,'' in \emph{Mobile Health}.\hskip 1em plus 0.5em minus
  0.4em\relax Springer, 2017, pp. 495--517.

\bibitem{nahum2018just}
I.~Nahum-Shani, S.~N. Smith, B.~J. Spring, L.~M. Collins, K.~Witkiewitz,
  A.~Tewari, and S.~A. Murphy, ``Just-in-time adaptive interventions (jitais)
  in mobile health: key components and design principles for ongoing health
  behavior support,'' \emph{Annals of Behavioral Medicine}, vol.~52, no.~6, pp.
  446--462, 2018.

\bibitem{lai1985asymptotically}
T.~L. Lai and H.~Robbins, ``Asymptotically efficient adaptive allocation
  rules,'' \emph{Advances in applied mathematics}, vol.~6, no.~1, pp. 4--22,
  1985.

\bibitem{abe1999associative}
N.~Abe and P.~M. Long, ``Associative reinforcement learning using linear
  probabilistic concepts,'' in \emph{ICML}.\hskip 1em plus 0.5em minus
  0.4em\relax Citeseer, 1999, pp. 3--11.

\bibitem{auer2002using}
P.~Auer, ``Using confidence bounds for exploitation-exploration trade-offs,''
  \emph{Journal of Machine Learning Research}, vol.~3, no. Nov, pp. 397--422,
  2002.

\bibitem{abbasi2011improved}
Y.~Abbasi-Yadkori, D.~P{\'a}l, and C.~Szepesv{\'a}ri, ``Improved algorithms for
  linear stochastic bandits,'' \emph{Advances in neural information processing
  systems}, vol.~24, pp. 2312--2320, 2011.

\bibitem{chu2011contextual}
W.~Chu, L.~Li, L.~Reyzin, and R.~Schapire, ``Contextual bandits with linear
  payoff functions,'' in \emph{Proceedings of the Fourteenth International
  Conference on Artificial Intelligence and Statistics}.\hskip 1em plus 0.5em
  minus 0.4em\relax JMLR Workshop and Conference Proceedings, 2011, pp.
  208--214.

\bibitem{faradonbeh2020optimism}
M.~K.~S. Faradonbeh, A.~Tewari, and G.~Michailidis, ``Optimism-based adaptive
  regulation of linear-quadratic systems,'' \emph{IEEE Transactions on
  Automatic Control}, vol.~66, no.~4, pp. 1802--1808, 2020.

\bibitem{thompson1933likelihood}
W.~R. Thompson, ``On the likelihood that one unknown probability exceeds
  another in view of the evidence of two samples,'' \emph{Biometrika}, vol.~25,
  no. 3/4, pp. 285--294, 1933.

\bibitem{strens2000bayesian}
M.~Strens, ``A bayesian framework for reinforcement learning,'' in \emph{ICML},
  vol. 2000, 2000, pp. 943--950.

\bibitem{chapelle2011empirical}
O.~Chapelle and L.~Li, ``An empirical evaluation of thompson sampling,''
  \emph{Advances in neural information processing systems}, vol.~24, pp.
  2249--2257, 2011.

\bibitem{kaufmann2012thompson}
E.~Kaufmann, N.~Korda, and R.~Munos, ``Thompson sampling: An asymptotically
  optimal finite-time analysis,'' in \emph{International conference on
  algorithmic learning theory}.\hskip 1em plus 0.5em minus 0.4em\relax
  Springer, 2012, pp. 199--213.

\bibitem{agrawal2012analysis}
S.~Agrawal and N.~Goyal, ``Analysis of thompson sampling for the multi-armed
  bandit problem,'' in \emph{Conference on learning theory}.\hskip 1em plus
  0.5em minus 0.4em\relax JMLR Workshop and Conference Proceedings, 2012, pp.
  39--1.

\bibitem{agrawal2013thompson}
------, ``Thompson sampling for contextual bandits with linear payoffs,'' in
  \emph{International Conference on Machine Learning}.\hskip 1em plus 0.5em
  minus 0.4em\relax PMLR, 2013, pp. 127--135.

\bibitem{faradonbeh2020adaptive}
M.~K.~S. Faradonbeh, A.~Tewari, and G.~Michailidis, ``On adaptive
  linear--quadratic regulators,'' \emph{Automatica}, vol. 117, p. 108982, 2020.

\bibitem{hu2019note}
T.~Hu, E.~B. Laber, Z.~Li, N.~J. Meyer, and K.~Pacifici, ``Note on thompson
  sampling for large decision problems,'' \emph{arXiv preprint
  arXiv:1905.04735}, 2019.

\bibitem{raghavan2020greedy}
M.~Raghavan, A.~Slivkins, J.~W. Vaughan, and Z.~S. Wu, ``Greedy algorithm
  almost dominates in smoothed contextual bandits,'' \emph{arXiv preprint
  arXiv:2005.10624}, 2020.

\bibitem{bastani2021mostly}
H.~Bastani, M.~Bayati, and K.~Khosravi, ``Mostly exploration-free algorithms
  for contextual bandits,'' \emph{Management Science}, vol.~67, no.~3, pp.
  1329--1349, 2021.

\bibitem{raykar2014sequential}
V.~Raykar and P.~Agrawal, ``Sequential crowdsourced labeling as an
  epsilon-greedy exploration in a markov decision process,'' in
  \emph{Artificial intelligence and statistics}.\hskip 1em plus 0.5em minus
  0.4em\relax PMLR, 2014, pp. 832--840.

\bibitem{gopalan2015thompson}
A.~Gopalan and S.~Mannor, ``Thompson sampling for learning parameterized markov
  decision processes,'' in \emph{Conference on Learning Theory}.\hskip 1em plus
  0.5em minus 0.4em\relax PMLR, 2015, pp. 861--898.

\bibitem{ouyang2017learning}
Y.~Ouyang, M.~Gagrani, A.~Nayyar, and R.~Jain, ``Learning unknown markov
  decision processes: A thompson sampling approach,'' \emph{arXiv preprint
  arXiv:1709.04570}, 2017.

\bibitem{abbasi2011regret}
Y.~Abbasi-Yadkori and C.~Szepesv{\'a}ri, ``Regret bounds for the adaptive
  control of linear quadratic systems,'' in \emph{Proceedings of the 24th
  Annual Conference on Learning Theory}.\hskip 1em plus 0.5em minus 0.4em\relax
  JMLR Workshop and Conference Proceedings, 2011, pp. 1--26.

\bibitem{faradonbeh2019applications}
M.~K.~S. Faradonbeh, A.~Tewari, and G.~Michailidis, ``On applications of
  bootstrap in continuous space reinforcement learning,'' in \emph{2019 IEEE
  58th Conference on Decision and Control (CDC)}.\hskip 1em plus 0.5em minus
  0.4em\relax IEEE, 2019, pp. 1977--1984.

\bibitem{faradonbeh2020input}
------, ``Input perturbations for adaptive control and learning,''
  \emph{Automatica}, vol. 117, p. 108950, 2020.

\bibitem{faradonbeh2021efficient}
M.~K.~S. Faradonbeh and M.~S.~S. Faradonbeh, ``Efficient estimation and control
  of unknown stochastic differential equations,'' \emph{arXiv preprint
  arXiv:2109.07630}, 2021.

\bibitem{kaelbling1998planning}
L.~P. Kaelbling, M.~L. Littman, and A.~R. Cassandra, ``Planning and acting in
  partially observable stochastic domains,'' \emph{Artificial intelligence},
  vol. 101, no. 1-2, pp. 99--134, 1998.

\bibitem{spaan2012partially}
M.~T. Spaan, ``Partially observable markov decision processes,'' in
  \emph{Reinforcement Learning}.\hskip 1em plus 0.5em minus 0.4em\relax
  Springer, 2012, pp. 387--414.

\bibitem{bensoussan2004stochastic}
A.~Bensoussan, \emph{Stochastic control of partially observable systems}.\hskip
  1em plus 0.5em minus 0.4em\relax Cambridge University Press, 2004.

\bibitem{roesser1975discrete}
R.~Roesser, ``A discrete state-space model for linear image processing,''
  \emph{IEEE Transactions on Automatic Control}, vol.~20, no.~1, pp. 1--10,
  1975.

\bibitem{nagrath2006control}
I.~Nagrath, \emph{Control systems engineering}.\hskip 1em plus 0.5em minus
  0.4em\relax New Age International, 2006.

\bibitem{kalman1960new}
R.~E. Kalman, ``A new approach to linear filtering and prediction problems,''
  1960.

\bibitem{stratonovich1960application}
R.~L. Stratonovich, ``Application of the markov processes theory to optimal
  filtering,'' \emph{Radio Engineering and Electronic Physics}, vol.~5, pp.
  1--19, 1960.

\bibitem{bouneffouf2017context}
D.~Bouneffouf, I.~Rish, G.~A. Cecchi, and R.~F{\'e}raud, ``Context attentive
  bandits: Contextual bandit with restricted context,'' \emph{arXiv preprint
  arXiv:1705.03821}, 2017.

\bibitem{park2021analysis}
H.~Park and M.~K.~S. Faradonbeh, ``Analysis of thompson sampling for partially
  observable contextual multi-armed bandits,'' \emph{IEEE Control Systems
  Letters}, 2021.

\bibitem{tennenholtz2021bandits}
G.~Tennenholtz, U.~Shalit, S.~Mannor, and Y.~Efroni, ``Bandits with partially
  observable confounded data,'' in \emph{Conference on Uncertainty in
  Artificial Intelligence. PMLR}, 2021.

\end{thebibliography}
\end{document}